\documentclass[letterpaper]{article} 
\usepackage[preprint]{aaai2027}  
\usepackage[hyphens]{url}  
\usepackage{graphicx} 
\usepackage{natbib}  
\usepackage{caption} 
\usepackage{amsmath}
\usepackage{amssymb}
\usepackage{booktabs}
\usepackage{multirow}
\usepackage[table]{xcolor}
\newcommand{\gaincell}[1]{\underline{#1}}

\title{Backbone-Agnostic Stochastic Perturbation Learning for \\End-to-End Real-World Image Dehazing}
\author{Bingcai Wei}
\affiliations{
School of Computer Science\\
Wuhan University\\
Wuhan, Hubei, China\\
\texttt{weibc97@whu.edu.cn}
}

\author{Yuning Cui}
\affiliations{
TUM School of Computation\\
Information and Technology\\
Technical University of Munich\\
\texttt{weibc97@whu.edu.cn}
}

\author{Mingyu Liu}
\affiliations{
CIT\\
Technische Universität München\\
\texttt{weibc97@whu.edu.cn}
}

\author{Jinni Geng}
\affiliations{
School of Optoelectronic Engineering\\
Xidian University\\
\texttt{weibc97@whu.edu.cn}
}

\author{Ling Li}
\affiliations{
computer and science\\
Tsinghua University\\
\texttt{weibc97@whu.edu.cn}
}

\author{Benwang Chen}
\affiliations{
Department of Automation\\
Tsinghua University\\
\texttt{weibc97@whu.edu.cn}
}

\author{Ziwei Li}
\affiliations{
King Abdullah University of Science and Technology\\
\texttt{weibc97@whu.edu.cn}
}

\author{Alois Knoll}
\affiliations{
Technical University of Munich\\
\texttt{weibc97@whu.edu.cn}
}

\author{
Bingcai Wei,\textsuperscript{\rm 1}
Yuning Cui,\textsuperscript{\rm 2}
Mingyu Liu,\textsuperscript{\rm 2}
Jinni Geng,\textsuperscript{\rm 3}
Ling Li,\textsuperscript{\rm 4}
Benwang Chen,\textsuperscript{\rm 4}
Ziwei Li,\textsuperscript{\rm 5}
Alois Knoll\textsuperscript{\rm 2}
}

\affiliations{
\textsuperscript{\rm 1}School of Computer Science, Wuhan University,
\textsuperscript{\rm 2}Technical University of Munich,
\textsuperscript{\rm 3}Xidian University,
\textsuperscript{\rm 4}Tsinghua University,
\textsuperscript{\rm 5}King Abdullah University of Science and Technology

weibc97@whu.edu.cn,
yuning.cui@in.tum.de,
mingyu.liu@tum.de,
22191110658@stu.xidian.edu.cn,
liling25@mails.tsinghua.edu.cn.
paper595236@163.com,
gslgsl00@126.com,
knoll@in.tum.de
}

\begin{document}

\maketitle

\begin{abstract}
Real-world paired image dehazing remains challenging because haze degradation is spatially non-uniform, illumination-dependent, and physically ambiguous even when haze-free references are available. Existing end-to-end restoration networks usually learn a deterministic mapping from a hazy observation to a clean target, while degradation-sensitive feature responses, physics-inspired reverse reconstruction, and cross-domain negative structure remain insufficiently exploited. In this paper, we propose Backbone-Agnostic Stochastic Perturbation Learning (BSPL), a plug-and-play framework for end-to-end real-world image dehazing. BSPL first introduces a Learnable Stochastic Perturbation Modulator (LSPM), which learns input-conditioned channel-wise and spatial-wise perturbation distributions and converts the resulting feature-response discrepancies into adaptive modulation weights. It then develops a Prior-informed Perturbation-guided Reconstruction Module (PPRM), which reuses the learned bottleneck perturbations together with transmission and atmospheric-light priors to reconstruct the hazy observation from the restored result and enforce physics-inspired degradation consistency. Furthermore, we propose a Dual-space Domain-diversified Distribution-aware Contrastive Loss ($D^3$CL) to regularize both clean restoration and hazy reconstruction spaces with real-world and synthetic negatives. Experiments on five real-world paired benchmarks show that BSPL consistently improves multiple representative backbones with only marginal additional inference overhead.
\end{abstract}


\section{Introduction}

Image dehazing aims to recover a visually faithful haze-free image from a hazy observation and remains a fundamental ill-posed low-level vision problem. The classical atmospheric scattering model describes a hazy image as
\begin{equation}
    I(x) = J(x)t(x) + A(1-t(x)),
\end{equation}
where $I(x)$ and $J(x)$ are the hazy observation and haze-free radiance, $t(x)$ is the transmission map, and $A$ is the global atmospheric light. Although this formulation motivates interpretable priors such as the dark channel prior \cite{he2011dark}, real haze rarely follows a deterministic pattern. Spatially varying density, illumination, depth ambiguity, camera response, and prior-estimation errors make the inverse mapping highly ambiguous and cause degradation responses to vary strongly across channels and spatial locations.

Real-world paired datasets, including I-HAZE \cite{ancuti2018ihaze}, O-HAZE \cite{ancuti2018ohaze}, Dense-Haze \cite{ancuti2019densehaze}, NH-HAZE \cite{ancuti2020nhhaze}, and LMHaze \cite{zhang2024lmhaze}, enable quantitative evaluation with aligned hazy and haze-free pairs and reveal the limitations of methods under diverse real degradations. Dense haze suppresses texture and contrast, non-homogeneous haze varies spatially, and multi-intensity haze creates strong cross-scene shifts. These variations make deterministic restoration learned mainly from paired reconstruction vulnerable to over-smoothed structures or color deviations despite visually plausible outputs.

Image restoration has progressed through stronger CNN, Transformer, and Mamba backbones \cite{cho2021mimo,chen2022nafnet,zamir2022restormer,wang2022uformer,cui2024oknet,li2025mair,wu2026c2ssm,wu2026hogformer,cui2026vivnet}, while dehazing methods increasingly exploit multi-scale feedback, color guidance, contrastive regularization, and uncertainty-aware processing \cite{dong2020msbdn,wu2021aecrnet,zheng2023curricular,hong2022udn,fang2025sgdn,liu2025funet}. Nevertheless, most end-to-end systems learn a deterministic mapping $\hat{J}=f_{\theta}(I)$ dominated by restoration error. They do not explicitly probe which channels and locations respond strongly to plausible feature perturbations, test whether the restored image explains the observed haze under physics-inspired priors, or separate clean targets from real and synthetic degradation domains in feature space. Existing methods often treat noise perturbations as disturbances to suppress; however, such small disturbances are ubiquitous in practical imaging and feature computation, and their response patterns can reveal degradation-sensitive channels and locations that are useful for adaptive restoration.

Motivated by this observation, we propose \textbf{Backbone-Agnostic Stochastic Perturbation Learning} (\textbf{BSPL}) with three complementary components. The \textbf{Learnable Stochastic Perturbation Modulator} (\textbf{LSPM}) predicts channel-wise and spatial-wise perturbation distributions, uses reparameterized sampling and inverse alignment to derive response discrepancies, and converts them into feature modulation weights. The \textbf{Prior-informed Perturbation-guided Reconstruction Module} (\textbf{PPRM}) reuses bottleneck perturbations and transmission/atmospheric-light priors to reconstruct the hazy observation from the restored output, encouraging both clean appearance and degradation consistency. The \textbf{Dual-space Domain-diversified Distribution-aware Contrastive Loss} ($\mathbf{D^3}$\textbf{CL}) regularizes both clean restoration and hazy reconstruction spaces using real and synthetic negatives together with stochastic perceptual features.

BSPL is backbone-agnostic and inference-efficient: LSPM can be inserted into CNN-, Transformer-, and Mamba-based encoder--decoder networks, whereas PPRM and $D^3$CL act only during training. Inference retains only the lightweight LSPM components, yielding consistent gains across representative backbones with marginal additional complexity.

Our contributions are summarized as follows:
\begin{itemize}
    \item We introduce BSPL, a backbone-agnostic stochastic perturbation learning framework that strengthens end-to-end real-world image dehazing through perturbation-response modulation, prior-informed reconstruction, and dual-space contrastive regularization.
    \item We propose LSPM to learn input-conditioned channel-wise and spatial-wise perturbation distributions and transform their response discrepancies into adaptive feature modulation without redesigning the underlying restoration backbone.
    \item We design PPRM to reuse LSPM perturbations together with transmission and atmospheric-light priors, reconstructing the hazy observation and imposing a physics-inspired degradation-consistency constraint.
    \item We develop $D^3$CL, a dual-space domain-diversified distribution-aware contrastive loss that regularizes clean restoration and hazy reconstruction with real-world and synthetic negatives in a stochastic perceptual feature space.
    \item Extensive experiments, ablation studies, and mechanism analysis on multiple real-world paired dehazing benchmarks demonstrate the effectiveness and generality of BSPL.
\end{itemize}

\section{Related Work}

\subsection{Real-World Image Dehazing}
Single-image dehazing recovers latent scene radiance from a hazy observation. Classical methods estimate transmission and atmospheric light under the atmospheric scattering model; the dark channel prior is a representative hand-crafted approach \cite{he2011dark}. Deep methods replace manual estimation with trainable networks: DehazeNet predicts transmission \cite{cai2016dehazenet}, AOD-Net reformulates dehazing as an all-in-one mapping \cite{li2017aodnet}, and DCPDN jointly estimates physical variables and the clean image \cite{zhang2018dcpdn}. Later architectures improve representation with multi-scale feedback, attention, and Transformers, including GridDehazeNet \cite{liu2019griddehazenet}, MSBDN \cite{dong2020msbdn}, FFA-Net \cite{qin2020ffa}, and DehazeFormer \cite{song2023dehazeformer}. Recent real-image methods further exploit codebook priors in RIDCP \cite{wu2023ridcp} and YCbCr-guided structural and color cues in SGDN \cite{fang2025sgdn}. Whereas most approaches learn deterministic restoration or prior-guided predictions, BSPL learns input-conditioned stochastic perturbations to probe degradation-sensitive feature responses and uses the resulting discrepancies for feature modulation, physics-inspired reconstruction consistency, and contrastive regularization.

\subsection{Image Restoration}
General image restoration has advanced through efficient CNNs, Transformers, and Mamba-based state-space models. Representative backbones include MIMOUNet \cite{cho2021mimo}, NAFNet \cite{chen2022nafnet}, Restormer and Uformer \cite{zamir2022restormer,wang2022uformer}, prompt-conditioned all-in-one restoration \cite{potlapalli2023promptir}, IRNeXt \cite{cui2023irnext}, OKNet \cite{cui2024oknet}, MaIR \cite{li2025mair}, $\mathrm{C}^2$SSM \cite{wu2026c2ssm}, HOGformer \cite{wu2026hogformer}, and VIVNet \cite{cui2026vivnet}. Although these architectures provide strong deterministic restoration capacity, they do not explicitly probe perturbation-sensitive feature behavior or enforce reverse haze-formation consistency. BSPL complements them by learning input-conditioned perturbations that expose degradation-sensitive responses and jointly support adaptive modulation, reverse haze consistency, and dual-space contrastive regularization.

\section{Method}

\subsection{Overview}
Given a real-world hazy image $I\in\mathbb{R}^{3\times H\times W}$ and its paired haze-free image $J$, our goal is to learn an end-to-end dehazing model that predicts a restored image $\hat{J}=F_{\theta}(I)$ while explicitly exploiting degradation-sensitive feature responses. Instead of designing a new restoration architecture, BSPL is a plug-and-play stochastic perturbation learning framework that preserves the primary blocks and hierarchical topology of the selected backbone.

\begin{figure}[t]
    \centering
    \includegraphics[width=0.98\linewidth]{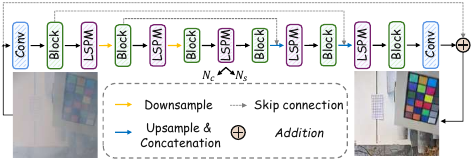}
    \caption{Overall integration of BSPL into a hierarchical encoder--decoder restoration network. The green Block modules denote the backbone network into which BSPL is embedded; LSPM units are interleaved with these original blocks without changing the backbone topology.}
    \label{fig:overall}
\end{figure}

As illustrated in Figure~\ref{fig:overall}, LSPM units modulate intermediate features at multiple resolutions without replacing the backbone blocks. The encoder progressively extracts degradation-aware representations, the decoder restores spatial detail through the original skip connections, and the final convolution predicts a residual that is added to the hazy input. This design allows the same stochastic perturbation learning principle to be embedded into CNN, Transformer, Mamba, or hybrid restoration backbones with minimal architectural intervention.

\begin{figure*}[t]
    \centering
    \includegraphics[width=0.90\linewidth]{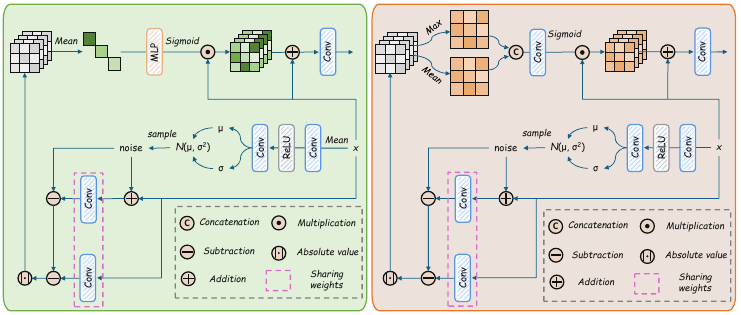}
    \caption{Illustration of LSPM. It estimates channel-wise and spatial-wise perturbation distributions, measures feature-response discrepancies under sampled perturbations, and converts them into modulation weights for feature enhancement.}
    \label{fig:lspm}
\end{figure*}

\begin{figure}[t]
    \centering
    \includegraphics[width=0.90\linewidth]{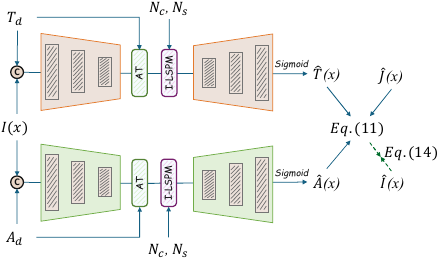}
    \caption{Illustration of PPRM. It uses transmission and atmospheric-light priors together to reconstruct the hazy observation through a physics-inspired formulation. AT denotes an affine transformation. I-LSPM denotes an inherited LSPM that receives $N_c$ and $N_s$ from the backbone bottleneck rather than learning independent perturbations.}
    \label{fig:pprm}
\end{figure}

\begin{figure}[t]
    \centering
    \includegraphics[width=0.90\linewidth]{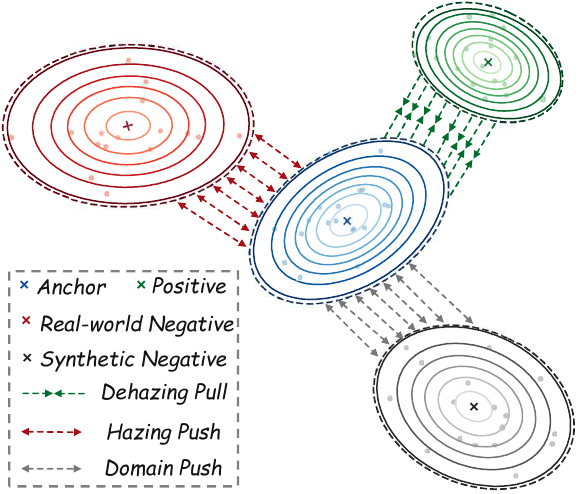}
    \caption{Illustration of $D^3$CL. Contrastive regularization is performed in both clean restoration and hazy reconstruction spaces, with real-world and synthetic negatives forming diversified degradation domains.}
    \label{fig:d3cl}
\end{figure}

The complete training framework contains three complementary components, detailed in Figs.~\ref{fig:lspm}--\ref{fig:d3cl}. LSPM produces perturbation-response-modulated features and sampled perturbation cues inside the restoration backbone. PPRM then uses the restored result $\hat{J}$, the hazy input $I$, dark-channel-prior estimates of transmission and atmospheric light, and the bottleneck perturbations shared from LSPM to reconstruct the predicted hazy image $\hat{I}$. Finally, $D^3$CL regularizes both the clean restoration space and the hazy reconstruction space with real-world and synthetic negatives.

\subsection{Learnable Stochastic Perturbation Modulator}
Neural feature tensors are generally sensitive to perturbations, but the response magnitude is input-dependent and varies across channels and spatial locations. This sensitivity is therefore not known a priori. LSPM exposes such heterogeneous responses and uses them as attention cues: features that exhibit a larger perturbation-induced discrepancy receive different modulation from features that remain comparatively stable. Let $X\in\mathbb{R}^{C\times H\times W}$ denote an intermediate feature map after layer normalization. Consider first a random channel-wise perturbation $N_c$ drawn from an input-conditioned Gaussian distribution. After adding $N_c$, applying a lightweight transformation, and subtracting the same perturbation to simulate an inverse-aligned perturbation response, the discrepancy
$D_c=|\psi_c(X)-(\psi_c(X+N_c)-N_c)|$
reflects how strongly the transformed feature responds to that perturbation. 
Specifically, global feature statistics are used to predict the distribution mean and log standard deviation:
\begin{equation}
    (\mu_c, \log \sigma_c) = f_c(\mathrm{GAP}(X)),
\end{equation}
where $\mu_c,\log \sigma_c\in\mathbb{R}^{C\times 1\times 1}$. We define the positive standard deviation as $\sigma_c=\exp(\log \sigma_c)>0$. The corresponding channel perturbation therefore follows $N_c\sim\mathcal{N}(\mu_c,\sigma_c^2)$ and is obtained by reparameterized sampling:
\begin{equation}
    N_c = \mu_c + \sigma_c \odot \epsilon_c, \quad \epsilon_c\sim\mathcal{N}(0, I).
\end{equation}
This parameterization permits end-to-end gradient propagation while guaranteeing a valid scale. Here $\mu_c$ determines the input-adaptive perturbation center, while $\sigma_c$ controls its stochastic spread. Within this conditional perturbation distribution, $\sigma_c$ describes the sampling uncertainty around $\mu_c$. After broadcasting $N_c$ over the spatial dimensions, the channel-wise sensitivity is explicitly computed as
\begin{equation}
    D_c = \left|\psi_c(X)-\left(\psi_c(X+N_c)-N_c\right)\right|,
\end{equation}
where $\psi_c(\cdot)$ denotes a lightweight convolutional transformation. A large value in $D_c$ indicates a strong perturbation response for the corresponding channel. LSPM then maps this sensitivity cue to a channel attention vector:
\begin{equation}
    W_c = \operatorname{Sigmoid}\left(g_c(\mathrm{GAP}(D_c))\right).
\end{equation}

The spatial-wise branch follows the same principle, but learns a location-dependent perturbation distribution from local features:
\begin{equation}
    (\mu_s, \log \sigma_s) = f_s(X),
\end{equation}
where $\mu_s,\log \sigma_s\in\mathbb{R}^{1\times H\times W}$ and $\sigma_s=\exp(\log \sigma_s)>0$. The spatial perturbation follows $N_s\sim\mathcal{N}(\mu_s,\sigma_s^2)$ and is sampled as
\begin{equation}
    N_s = \mu_s + \sigma_s \odot \epsilon_s, \quad \epsilon_s\sim\mathcal{N}(0, I),
\end{equation}
then replicated across channels and injected into $X$. The resulting spatial sensitivity map is
\begin{equation}
    D_s = \left|\psi_s(X)-\left(\psi_s(X+N_s)-N_s\right)\right|.
\end{equation}
LSPM aggregates this response discrepancy with average and maximum statistics along the channel dimension and produces a spatial attention map:
\begin{equation}
    W_s = \operatorname{Sigmoid}\left(g_s\left([\mathrm{Avg}_c(D_s), \mathrm{Max}_c(D_s)]\right)\right).
\end{equation}
The channel- and spatial-modulated features are fused with branch-wise residual compensation and a final residual connection:
\begin{equation}
\begin{aligned}
    X_c &= X\odot W_c + X, \quad X_s = X\odot W_s + X, \\
    \tilde{X} &= X + \eta_c(X_c) + \eta_s(X_s),
\end{aligned}
\end{equation}
where $\eta_c(\cdot)$ and $\eta_s(\cdot)$ are lightweight convolutional projections. Thus, the sampled noise is not used directly as an attention representation. Instead, LSPM learns input-conditioned random perturbations, measures the feature sensitivity they reveal, and converts the resulting channel-wise and spatial-wise discrepancies into adaptive attention weights. The sampled bottleneck perturbations are also passed to PPRM to couple forward dehazing and reverse haze reconstruction.

\subsection{Prior-informed Perturbation-guided Reconstruction Module}
Pixel-wise supervision encourages $\hat{J}$ to approach the haze-free target $J$, but it does not guarantee that the prediction remains consistent with the observed haze degradation under a physics-inspired formation view. To impose such a training constraint, PPRM reconstructs the hazy observation from the restored result through a parameterization inspired by the atmospheric scattering model:
\begin{equation}
    \hat{I}(x)=\hat{T}(x)\hat{J}(x)+(1-\hat{T}(x))\hat{A}(x),
    \label{eq:pprm_rec}
\end{equation}
where $\hat{T}(x)$ and $\hat{A}(x)$ are transmission-related and atmospheric-light-related maps predicted by PPRM. Equation~\eqref{eq:pprm_rec} is used as a physics-inspired reconstruction form rather than a strict physical inversion: because both maps are optimized end-to-end for reconstruction, $\hat{T}$ and $\hat{A}$ are not required to equal the true scene transmission and atmospheric light or to possess exact physical meanings. 

PPRM contains two prior-informed reconstruction branches. The transmission branch takes the hazy image and a DCP-estimated transmission prior $T_d$ as input, while the atmospheric-light branch takes the hazy image and a DCP-estimated atmospheric-light prior $A_d$ as input. These priors are injected in two ways. First, they are concatenated with the image input at multiple scales so that the reconstruction branch can exploit prior-aware local features. Second, they generate affine modulation parameters at the latent layer:
\begin{equation}
    Z_T' = Z_T \odot M_T(T_d) + B_T(T_d),
\end{equation}
\begin{equation}
    Z_A' = Z_A \odot M_A(A_d) + B_A(A_d),
\end{equation}
where $Z_T$ and $Z_A$ denote latent representations in the transmission and atmospheric-light branches. The outputs $M_T(T_d)$, $B_T(T_d)$, $M_A(A_d)$, and $B_A(A_d)$ are prior-conditioned features mapped by convolutional layers to provide multiplicative and additive affine parameters. This guidance regulates latent reconstruction features beyond shallow input concatenation.

To connect PPRM with the forward stochastic perturbation pathway, the I-LSPM blocks in Figure~\ref{fig:pprm} inherit $N_c$ and $N_s$ from the LSPM at the backbone bottleneck rather than estimating independent perturbations. Each reconstruction branch evaluates its own response to these shared perturbations and maps the response discrepancy to latent modulation weights. Consequently, channels or locations that react strongly in the forward dehazing pathway provide aligned perturbation cues for reverse haze reconstruction, while the transmission and atmospheric-light branches remain free to learn branch-specific modulation. The resulting reconstruction loss is defined as
\begin{equation}
    \mathcal{L}_{rec}=\rho(\hat{I}-I),
\end{equation}
where $\rho(\cdot)$ denotes the Charbonnier penalty. 

\subsection{Dual-space Domain-diversified Distribution-aware Contrastive Loss}
Conventional dehazing contrastive learning pulls the restored result toward the clean target and pushes it away from a negative hazy image in a deterministic perceptual feature space. However, real-world dehazing involves both clean restoration and haze reconstruction, and the negative domain may contain real-world haze and synthetic haze with different distributions. To address this issue, we propose $D^3$CL.

Let $a$, $p$, $n$, and $s$ denote an anchor, a positive sample, a real-world negative sample, and a synthetic negative sample, respectively. The real-world negative $n$ is a non-corresponding hazy image sampled from another training pair, rather than the hazy input paired with the current positive sample. Since the paired ground truth already provides content-preserving supervision, such dataset-level real-haze negatives allow the contrastive objective to model the broader real-world degradation domain and avoid collapsing into a redundant paired reconstruction constraint. A frozen VGG-19 network extracts multi-level perceptual features $\{\phi_l(\cdot)\}_{l=1}^{L}$. To approximate a distribution-aware feature representation without repeated network forwarding, we perturb each perceptual feature according to its activation magnitude:
\begin{equation}
    \tilde{\phi}_l(x)=\phi_l(x)+\epsilon_l\odot \sigma_l(x),
\end{equation}
\begin{equation}
    \sigma_l(x)=\sqrt{\frac{r}{1-r}}\,|\phi_l(x)|, \quad \epsilon_l\sim\mathcal{N}(0,I),
\end{equation}
where $r$ is a fixed perturbation ratio. This design provides an efficient approximation of stochastic perceptual feature distributions and avoids the high cost of multiple Monte Carlo VGG inferences.

The layer-wise distances are computed as
\begin{equation}
    d_l^{ap}=\left\|\tilde{\phi}_l(a)-\tilde{\phi}_l(p)\right\|_1,
\end{equation}
\begin{equation}
    d_l^{an}=\left\|\tilde{\phi}_l(a)-\tilde{\phi}_l(n)\right\|_1, \quad
    d_l^{as}=\left\|\tilde{\phi}_l(a)-\tilde{\phi}_l(s)\right\|_1.
\end{equation}
Then $D^3$CL is formulated as
\begin{equation}
    \mathcal{L}_{D^3\mathrm{CL}}(a,p,n,s)
    =\sum_{l=1}^{L}\lambda_l\frac{1}{2}
    \left(
    \frac{d_l^{ap}}{d_l^{an}+\varepsilon}
    +
    \frac{d_l^{ap}}{d_l^{as}+\varepsilon}
    \right),
    \label{eq:d3cl}
\end{equation}
where $\lambda_l$ balances different VGG layers and $\varepsilon$ avoids numerical instability. In our implementation, we use five VGG feature levels with $\{\lambda_l\}_{l=1}^{5}=\{1/32,1/16,1/8,1/4,1\}$ and set $r=0.2$. The two denominator terms correspond to the real-world degradation push and the synthetic-domain push, respectively.

We apply $D^3$CL in two spaces. In the clean restoration space, the dehazed output $\hat{J}$ is treated as the anchor, the clean image $J$ is the positive sample, and real-world and synthetic negative samples are used to enlarge the degradation margin:
\begin{equation}
    \mathcal{L}_{clean}^{D^3\mathrm{CL}}=\mathcal{L}_{D^3\mathrm{CL}}(\hat{J},J,N_r,N_s).
\end{equation}
In the hazy reconstruction space, the reconstructed hazy image $\hat{I}$ is treated as the anchor and the original hazy image $I$ is the positive sample:
\begin{equation}
    \mathcal{L}_{hazy}^{D^3\mathrm{CL}}=\mathcal{L}_{D^3\mathrm{CL}}(\hat{I},I,N_r^h,N_s^h).
\end{equation}
Here $N_s$ is sampled from the synthetic Haze4K domain with different image content. For the hazy reconstruction space, $N_r^h$ and $N_s^h$ are non-corresponding clean images from the real training set and Haze4K, respectively. This dual-space formulation simultaneously constrains the forward dehazing output and the backward haze reconstruction, making the contrastive regularization consistent with the bidirectional design of BSPL.

\subsection{Training Objective and Inference}
The overall training objective combines restoration supervision, haze reconstruction supervision, and dual-space contrastive regularization:
\begin{equation}
\begin{aligned}
    \mathcal{L} =
    &\ \rho(\hat{J}-J)
    + \rho(\hat{I}-I) \\
    &+ \frac{1}{2}\mathcal{L}_{clean}^{D^3\mathrm{CL}}
    + \frac{1}{2}\mathcal{L}_{hazy}^{D^3\mathrm{CL}}.
\end{aligned}
\label{eq:total_loss}
\end{equation}
The first term trains the backbone to restore the haze-free image. The second term enforces physics-inspired reconstruction consistency with the observed haze rather than exact recovery of physical variables. The last two terms regularize the feature distributions in both clean and hazy spaces.

At inference time, only the restoration backbone equipped with LSPM is used:
\begin{equation}
    \hat{J}=F_{\theta}^{\mathrm{LSPM}}(I).
\end{equation}
PPRM is removed during inference, and the LSPM random seed is fixed for deterministic evaluation. Therefore, the additional inference cost of BSPL comes only from the lightweight LSPM inserted into the backbone, while the physics-inspired reconstruction benefits training without slowing down deployment.

\section{Experiments}

\subsection{Experimental Setup}
\textbf{Datasets.}
We evaluate BSPL on five real-world paired image dehazing benchmarks with distinct degradation characteristics: Dense-Haze \cite{ancuti2019densehaze}, I-HAZE \cite{ancuti2018ihaze}, O-HAZE \cite{ancuti2018ohaze}, NH-HAZE \cite{ancuti2020nhhaze}, and LMHaze \cite{zhang2024lmhaze}. Dense-Haze contains dense homogeneous haze, I-HAZE and O-HAZE focus on indoor and outdoor real haze, respectively, NH-HAZE introduces spatially non-homogeneous haze, and LMHaze covers large-scale scenes with multiple haze intensities. The main training set aggregates 4,065 paired images: 40 from Dense-Haze, 25 from I-HAZE, 35 from O-HAZE, 40 from NH-HAZE, and 3,925 from LMHaze. Evaluation uses 15, 5, 10, 15, and 1,115 test pairs from these datasets, respectively. We report peak signal-to-noise ratio (PSNR) and structural similarity index (SSIM); higher values indicate better restoration quality.
All baselines are retrained under the same pipeline for fair pairwise comparison, with implementation details provided in the supplementary material.

\begin{table*}[t]
\centering

\setlength{\tabcolsep}{3.2pt}
\renewcommand{\arraystretch}{1.05}

\resizebox{0.9\textwidth}{!}{%
\begin{tabular}{@{}lcccccccc@{}}
\toprule
Model
& Venue
& Setting
& Dense-Haze
& I-HAZE
& O-HAZE
& NH-HAZE
& LMHaze
& Average \\
\midrule

\multirow{3}{*}{MIMOUNet~\cite{cho2021mimo}}
& ICCV'21
& Baseline
& 15.49/0.576
& 18.80/0.813
& 21.50/0.861
& 16.69/0.674
& 18.46/0.752
& 18.19/0.735 \\
& Ours
& \textbf{+BSPL}
& \textbf{16.05/0.592}
& \textbf{20.81/0.848}
& \textbf{23.54/0.887}
& \textbf{17.21/0.709}
& \textbf{19.01/0.796}
& \textbf{19.32/0.767} \\
& --
& \underline{Gain}
& \gaincell{+0.56/+0.016}
& \gaincell{+2.01/+0.035}
& \gaincell{+2.04/+0.026}
& \gaincell{+0.52/+0.035}
& \gaincell{+0.55/+0.044}
& \gaincell{+1.13/+0.032} \\
\midrule

\multirow{3}{*}{IRNeXt~\cite{cui2023irnext}}
& ICML'23
& Baseline
& 16.51/0.599
& 17.33/0.786
& 23.58/0.875
& 16.43/0.647
& 18.09/0.747
& 18.39/0.731 \\
& Ours
& \textbf{+BSPL}
& \textbf{16.63/0.602}
& \textbf{19.93/0.818}
& \textbf{24.58/0.882}
& \textbf{17.18/0.679}
& \textbf{18.46/0.783}
& \textbf{19.36/0.753} \\
& --
& \underline{Gain}
& \gaincell{+0.12/+0.003}
& \gaincell{+2.60/+0.032}
& \gaincell{+1.00/+0.007}
& \gaincell{+0.75/+0.032}
& \gaincell{+0.37/+0.036}
& \gaincell{+0.97/+0.022} \\
\midrule

\multirow{3}{*}{$\mathrm{C}^{2}$SSM~\cite{wu2026c2ssm}}
& CVPR'26
& Baseline
& 13.39/0.420
& 16.34/0.720
& 17.78/0.771
& 13.36/0.542
& 16.91/0.645
& 15.56/0.619 \\
& Ours
& \textbf{+BSPL}
& \textbf{16.18/0.575}
& \textbf{18.47/0.811}
& \textbf{21.33/0.870}
& \textbf{16.24/0.655}
& \textbf{18.14/0.760}
& \textbf{18.07/0.734} \\
& --
& \underline{Gain}
& \gaincell{+2.79/+0.155}
& \gaincell{+2.13/+0.091}
& \gaincell{+3.55/+0.099}
& \gaincell{+2.88/+0.113}
& \gaincell{+1.23/+0.115}
& \gaincell{+2.51/+0.115} \\
\midrule

\multirow{3}{*}{HOGformer-S~\cite{wu2026hogformer}}
& AAAI'26
& Baseline
& 15.74/0.562
& 18.46/0.759
& 20.94/0.852
& 16.01/0.666
& 17.91/0.730
& 17.81/0.714 \\
& Ours
& \textbf{+BSPL}
& \textbf{16.81/0.619}
& \textbf{19.96/0.844}
& \textbf{23.49/0.892}
& \textbf{17.37/0.728}
& \textbf{18.88/0.792}
& \textbf{19.30/0.775} \\
& --
& \underline{Gain}
& \gaincell{+1.07/+0.057}
& \gaincell{+1.50/+0.085}
& \gaincell{+2.55/+0.040}
& \gaincell{+1.36/+0.062}
& \gaincell{+0.97/+0.062}
& \gaincell{+1.49/+0.061} \\
\midrule

\multirow{3}{*}{VIVNet~\cite{cui2026vivnet}}
& TPAMI'26
& Baseline
& 15.19/0.569
& 18.00/0.784
& 22.49/0.861
& 17.09/0.697
& 18.46/0.760
& 18.24/0.734 \\
& Ours
& \textbf{+BSPL}
& \textbf{16.65/0.600}
& \textbf{18.94/0.828}
& \textbf{24.15/0.895}
& \textbf{17.84/0.739}
& \textbf{19.12/0.799}
& \textbf{19.34/0.772} \\
& --
& \underline{Gain}
& \gaincell{+1.46/+0.031}
& \gaincell{+0.94/+0.044}
& \gaincell{+1.66/+0.034}
& \gaincell{+0.75/+0.042}
& \gaincell{+0.66/+0.039}
& \gaincell{+1.10/+0.038} \\

\bottomrule
\end{tabular}%
}
\caption{Pairwise quantitative comparison of the original backbones and their BSPL-enhanced variants on five real-world paired dehazing benchmarks. Each performance entry is reported as PSNR/SSIM. The ``Gain'' rows show the absolute improvements brought by BSPL over the corresponding baselines.}
\label{tab:main_results}
\end{table*}
\begin{figure*}[t]
\centering
\setlength{\tabcolsep}{0.7pt}
\resizebox{0.9\textwidth}{!}{%
\begin{tabular}{@{}cccccccccccc@{}}

\includegraphics[width=0.078\textwidth]{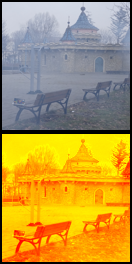} &
\includegraphics[width=0.078\textwidth]{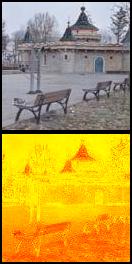} &
\includegraphics[width=0.078\textwidth]{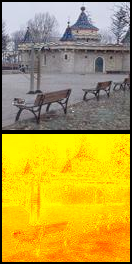} &
\includegraphics[width=0.078\textwidth]{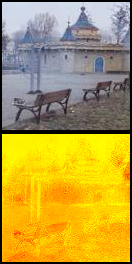} &
\includegraphics[width=0.078\textwidth]{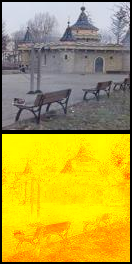} &
\includegraphics[width=0.078\textwidth]{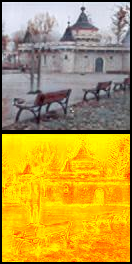} &
\includegraphics[width=0.078\textwidth]{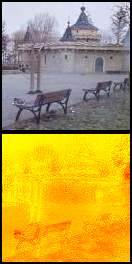} &
\includegraphics[width=0.078\textwidth]{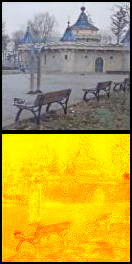} &
\includegraphics[width=0.078\textwidth]{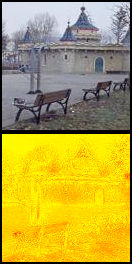} &
\includegraphics[width=0.078\textwidth]{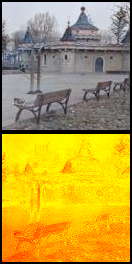} &
\includegraphics[width=0.078\textwidth]{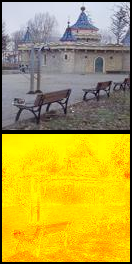} &
\includegraphics[width=0.078\textwidth]{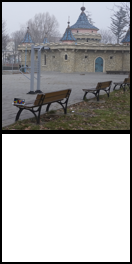} \\

\includegraphics[width=0.078\textwidth]{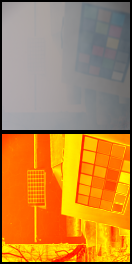} &
\includegraphics[width=0.078\textwidth]{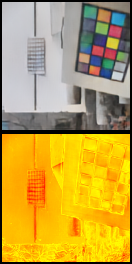} &
\includegraphics[width=0.078\textwidth]{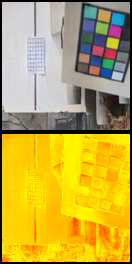} &
\includegraphics[width=0.078\textwidth]{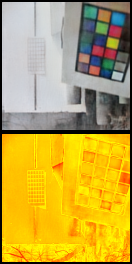} &
\includegraphics[width=0.078\textwidth]{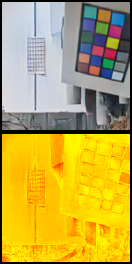} &
\includegraphics[width=0.078\textwidth]{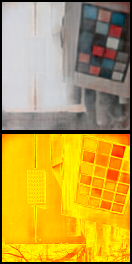} &
\includegraphics[width=0.078\textwidth]{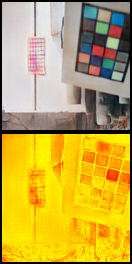} &
\includegraphics[width=0.078\textwidth]{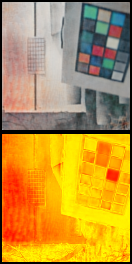} &
\includegraphics[width=0.078\textwidth]{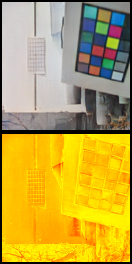} &
\includegraphics[width=0.078\textwidth]{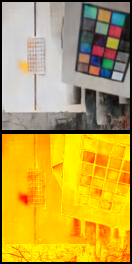} &
\includegraphics[width=0.078\textwidth]{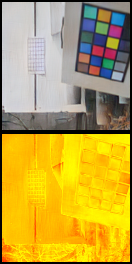} &
\includegraphics[width=0.078\textwidth]{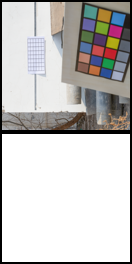} \\

(a) & (b) & (c) & (d) & (e) & (f) &
(g) & (h) & (i) & (j) & (k) & (l)

\end{tabular}%
}
\caption{Qualitative comparison.
(a) Input;
(b) MIMOUNet~\cite{cho2021mimo};
(c) MIMOUNet+BSPL;
(d) IRNeXt~\cite{cui2023irnext};
(e) IRNeXt+BSPL;
(f) $\mathrm{C}^{2}$SSM~\cite{wu2026c2ssm};
(g) $\mathrm{C}^{2}$SSM+BSPL;
(h) HOGformer-S~\cite{wu2026hogformer};
(i) HOGformer-S+BSPL;
(j) VIVNet~\cite{cui2026vivnet};
(k) VIVNet+BSPL;
and (l) ground truth.
Following~\cite{zheng2021wdc}, each panel stacks the RGB result (top) and its absolute error map relative to the ground truth (bottom); darker colors indicate larger errors.}
\label{fig:qualitative}
\end{figure*}
\subsection{Quantitative and Qualitative Comparisons}
Table~\ref{tab:main_results} presents pairwise comparisons between five representative backbones and their BSPL-enhanced variants. BSPL consistently improves all backbones across the five benchmarks, yielding average gains of $1.13$ dB/$0.032$ SSIM for MIMOUNet, $0.97$ dB/$0.022$ for IRNeXt, $2.51$ dB/$0.115$ for $\mathrm{C}^{2}$SSM, $1.49$ dB/$0.061$ for HOGformer-S, and $1.10$ dB/$0.038$ for VIVNet. These consistent gains across CNN-, Mamba-, Transformer-, and hybrid architectures demonstrate the backbone-agnostic effectiveness of BSPL. The gain rows further show that BSPL remains effective under diverse degradation conditions. In particular, $\mathrm{C}^{2}$SSM obtains the largest average improvement, including $3.55$ dB on O-HAZE and $2.88$ dB on NH-HAZE. Among the enhanced models, HOGformer-S+BSPL performs best on Dense-Haze, MIMOUNet+BSPL on I-HAZE, IRNeXt+BSPL on O-HAZE, and VIVNet+BSPL on NH-HAZE and LMHaze, indicating improved robustness to varying haze distributions.

Figure~\ref{fig:qualitative} visually confirms these improvements. Since subtle differences are difficult to distinguish from RGB results alone, we additionally present absolute error maps relative to ground truth following~\cite{zheng2021wdc}. Across both scenes, BSPL reduces residual haze and color distortion while preserving finer structures, with fewer dark high-error regions than the corresponding baselines.

\subsection{Ablation Studies}
\begin{table}[t]
\centering
\begin{tabular}{@{}p{0.60\columnwidth}cc@{}}
\toprule
Configuration & PSNR & SSIM \\
\midrule
Baseline & 17.14 & 0.698 \\
\midrule
w/o channel-wise branch & 17.40 & 0.716 \\
w/o spatial-wise branch & 17.43 & 0.719 \\
w/o learnable perturbation & 17.32 & 0.711 \\
Direct discrepancy weighting & 17.32 & 0.713 \\
\textbf{Full LSPM} & \textbf{17.56} & \textbf{0.725} \\
\midrule
w/o $T_d$ and $A_d$ input & 17.36 & 0.703 \\
w/o $T_d$ input & 17.39 & 0.718 \\
w/o $A_d$ input & 17.38 & 0.715 \\
w/o latent affine guidance & 17.53 & 0.725 \\
w/o perturbation-guided modulation & 17.68 & 0.729 \\
Independent perturbations & 17.85 & 0.741 \\
\textbf{Full PPRM (shared $N_c,N_s$)} & \textbf{18.03} & \textbf{0.753} \\
\bottomrule
\end{tabular}
\caption{Ablation of LSPM and PPRM.``Independent'' means that PPRM estimates its own perturbations rather than reusing $N_c$ and $N_s$ from the backbone LSPM.} 
\label{tab:ablation_modules}
\end{table}

\textbf{Effect of LSPM.}
The upper part of Table~\ref{tab:ablation_modules} validates the roles of LSPM. Removing either the channel-wise or spatial-wise branch degrades performance, indicating that degradation sensitivity is distributed across both dimensions. Disabling learnable perturbation generation reduces PSNR from $17.56$ to $17.32$ dB, demonstrating that input-conditioned perturbation distributions are important for probing useful response differences. Direct discrepancy weighting is also inferior to the complete design, confirming that raw perturbation responses should be adaptively mapped before feature modulation. Overall, LSPM improves the baseline by $0.42$ dB PSNR and $0.027$ SSIM.

\textbf{Effect of PPRM.}
The lower part of Table~\ref{tab:ablation_modules} progressively constructs PPRM. Removing both $T_d$ and $A_d$ causes the largest prior-input degradation, while removing either one also hurts performance, showing their complementarity. Latent affine guidance improves over shallow concatenation, and removing perturbation-guided modulation reduces the result to $17.68/0.729$. 
Independently estimating its perturbations lowers performance from $18.03/0.753$ to $17.85/0.741$, supporting their intended cross-module interaction.

\begin{table}[t]
\centering
\begin{tabular}{@{}p{0.60\columnwidth}cc@{}}
\toprule
Configuration & PSNR & SSIM \\
\midrule
$\mathcal{L}_{char}$ only & 17.14 & 0.698 \\
$+$ conventional CRL & 17.73 & 0.739 \\
$+$ non-corresp. real negative & 17.79 & 0.742 \\
$+$ synthetic negative domain & 17.84 & 0.754 \\
$+$ stochastic VGG features & 17.93 & 0.757 \\
Symmetric KL distance & 17.63 & 0.733 \\
$W_2$ distance & 17.64 & 0.736 \\
\textbf{Full $D^3$CL (clean+hazy)} & \textbf{18.23} & \textbf{0.767} \\
\bottomrule
\end{tabular}
\caption{Progressive construction of $D^3$CL. ``Non-corresp.'' denotes a real hazy negative sampled from another pair.}
\label{tab:ablation_d3cl}
\end{table}

\textbf{Effect of $D^3$CL.}
Table~\ref{tab:ablation_d3cl} progressively constructs $D^3$CL from $\mathcal{L}_{char}$. Conventional contrastive regularization improves the baseline, and a non-corresponding real hazy negative broadens the real-haze domain. Adding a synthetic negative raises SSIM from $0.742$ to $0.754$, confirming complementary negative domains. Stochastic VGG features reach $17.93/0.757$, while symmetric KL and Wasserstein-2 are less effective than sampled $\ell_1$ distance. 
Extending regularization to the reconstructed-hazy space yields $18.23/0.767$, reflecting the complete interaction of all three components.

\subsection{Mechanism Analysis}

\begin{center}
\begin{minipage}{\columnwidth}
\centering
\setlength{\tabcolsep}{0.35pt}
\resizebox{\textwidth}{!}{%
\begin{tabular}{@{}ccccc@{}}
\includegraphics[width=0.192\linewidth]{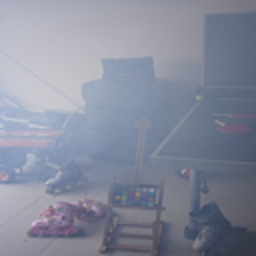} &
\includegraphics[width=0.192\linewidth]{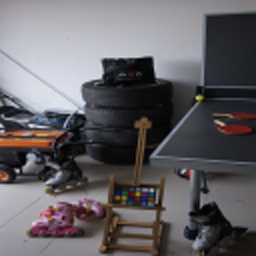} &
\includegraphics[width=0.192\linewidth]{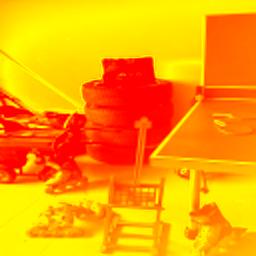} &
\includegraphics[width=0.192\linewidth]{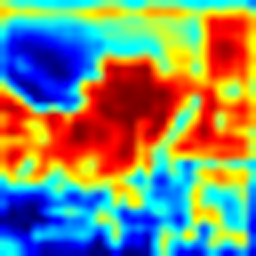} &
\includegraphics[width=0.192\linewidth]{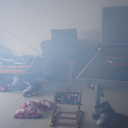} \\
\includegraphics[width=0.192\linewidth]{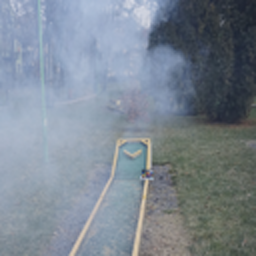} &
\includegraphics[width=0.192\linewidth]{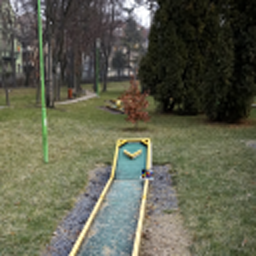} &
\includegraphics[width=0.192\linewidth]{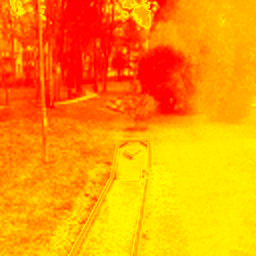} &
\includegraphics[width=0.192\linewidth]{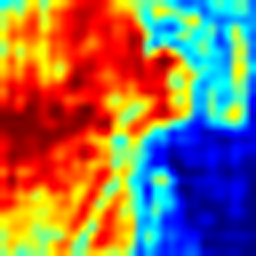} &
\includegraphics[width=0.192\linewidth]{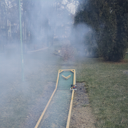} \\
(a) & (b) & (c) & (d) & (e)
\end{tabular}
}
\captionof{figure}{Perturbation-response and haze-reconstruction analysis. (a) Input, (b) ground truth, (c) absolute input--GT error map, (d) LSPM spatial perturbation-response map, and (e) PPRM-reconstructed hazy image. The two rows show representative real-world hazy scenes.}
\label{fig:perturbation_reconstruction}
\end{minipage}
\end{center}

Figure~\ref{fig:perturbation_reconstruction} compares pixel-space degradation with feature-space sensitivity. The input-GT error map visualizes spatial haze, whereas the LSPM map highlights locations with strong perturbation-response discrepancies; they need not coincide, measuring different representation spaces. PPRM reconstructs dominant haze appearance and distribution, supporting complementary forward modulation and reverse degradation consistency. Two examples further show LSPM does not simply reproduce pixel residuals: its responses emphasize representation regions associated with haze boundaries, texture attenuation, and non-uniform veiling. Meanwhile, PPRM preserves scene-dependent haze patterns rather than arbitrary perturbations, indicating shared cues remain relevant across forward restoration and reverse reconstruction paths. Together, these observations support the mechanism of using stochastic probes to expose degradation-sensitive representations and reconstruction consistency to align these cues with real haze formation.

\section{Conclusion}
We presented BSPL, a backbone-agnostic stochastic perturbation learning framework for end-to-end real-world image dehazing. BSPL uses LSPM to learn input-conditioned perturbation distributions and transform their response discrepancies into channel-wise and spatial-wise feature modulation, PPRM to impose physics-inspired reverse haze reconstruction with shared perturbations, and $D^3$CL to regularize clean and hazy feature distributions with diversified negative domains. Experiments on five real-world paired benchmarks demonstrate consistent improvements across five representative restoration backbones. These gains indicate that the learned perturbation-response cues are not tied to a particular feature operator or network family, supporting BSPL as a transferable enhancement rather than a backbone-specific redesign. These results show that probing degradation-sensitive feature responses with stochastic perturbations and enforcing physics-inspired degradation consistency provide a general and efficient way to strengthen real-world dehazing networks.


\bibliography{bspl_refs}

@article{he2011dark,
  title={Single Image Haze Removal Using Dark Channel Prior},
  author={He, Kaiming and Sun, Jian and Tang, Xiaoou},
  journal={IEEE Transactions on Pattern Analysis and Machine Intelligence},
  volume={33},
  number={12},
  pages={2341--2353},
  year={2011}
}

@inproceedings{ancuti2018ihaze,
  title={I-HAZE: A Dehazing Benchmark with Real Hazy and Haze-Free Indoor Images},
  author={Ancuti, Codruta O. and Ancuti, Cosmin and Timofte, Radu and De Vleeschouwer, Christophe},
  booktitle={Advanced Concepts for Intelligent Vision Systems},
  pages={620--631},
  year={2018}
}

@inproceedings{ancuti2018ohaze,
  title={O-HAZE: A Dehazing Benchmark with Real Hazy and Haze-Free Outdoor Images},
  author={Ancuti, Codruta O. and Ancuti, Cosmin and Timofte, Radu and De Vleeschouwer, Christophe},
  booktitle={Proceedings of the IEEE Conference on Computer Vision and Pattern Recognition Workshops},
  pages={754--762},
  year={2018}
}

@inproceedings{ancuti2019densehaze,
  title={Dense-Haze: A Benchmark for Image Dehazing with Dense-Haze and Haze-Free Images},
  author={Ancuti, Codruta O. and Ancuti, Cosmin and Sbert, Mateu and Timofte, Radu},
  booktitle={IEEE International Conference on Image Processing},
  pages={1014--1018},
  year={2019}
}

@inproceedings{ancuti2020nhhaze,
  title={NH-HAZE: An Image Dehazing Benchmark with Non-Homogeneous Hazy and Haze-Free Images},
  author={Ancuti, Codruta O. and Ancuti, Cosmin and Timofte, Radu},
  booktitle={Proceedings of the IEEE/CVF Conference on Computer Vision and Pattern Recognition Workshops},
  pages={444--445},
  year={2020}
}

@inproceedings{zhang2024lmhaze,
  title={LMHaze: Intensity-Aware Image Dehazing with a Large-Scale Multi-Intensity Real Haze Dataset},
  author={Zhang, Ruikun and Yang, Hao and Yang, Yan and Fu, Ying and Pan, Liyuan},
  booktitle={Proceedings of the ACM Multimedia Asia},
  year={2024}
}

@inproceedings{hong2022udn,
  title={Uncertainty-Driven Dehazing Network},
  author={Hong, Ming and Liu, Jinshan and Li, Cuihua and Qu, Yanyun},
  booktitle={Proceedings of the AAAI Conference on Artificial Intelligence},
  volume={36},
  number={1},
  pages={906--913},
  year={2022}
}

@inproceedings{wu2021aecrnet,
  title={Contrastive Learning for Compact Single Image Dehazing},
  author={Wu, Haiyan and Qu, Yanyun and Lin, Shaohui and Zhou, Jian and Qiao, Ruizhi and Zhang, Zhizhong and Xie, Yuan and Ma, Lizhuang},
  booktitle={Proceedings of the IEEE/CVF Conference on Computer Vision and Pattern Recognition},
  pages={10551--10560},
  year={2021}
}

@inproceedings{cho2021mimo,
  title={Rethinking Coarse-to-Fine Approach in Single Image Deblurring},
  author={Cho, Sung-Jin and Ji, Seo-Won and Hong, Jun-Pyo and Jung, Seung-Won and Ko, Sung-Jea},
  booktitle={Proceedings of the IEEE/CVF International Conference on Computer Vision},
  pages={4641--4650},
  year={2021}
}

@inproceedings{chen2022nafnet,
  title={Simple Baselines for Image Restoration},
  author={Chen, Liangyu and Chu, Xiaojie and Zhang, Xiangyu and Sun, Jian},
  booktitle={European Conference on Computer Vision},
  pages={17--33},
  year={2022}
}

@inproceedings{cui2024oknet,
  title={Omni-Kernel Network for Image Restoration},
  author={Cui, Yuning and Ren, Wenqi and Knoll, Alois},
  booktitle={Proceedings of the AAAI Conference on Artificial Intelligence},
  volume={38},
  number={2},
  pages={1426--1434},
  year={2024}
}

@inproceedings{fang2025sgdn,
  title={Guided Real Image Dehazing Using YCbCr Color Space},
  author={Fang, Wenxuan and Fan, Junkai and Zheng, Yu and Weng, Jiangwei and Tai, Ying and Li, Jun},
  booktitle={Proceedings of the AAAI Conference on Artificial Intelligence},
  volume={39},
  number={3},
  pages={2787--2795},
  year={2025}
}

@inproceedings{wu2026hogformer,
  title={Gradient as Conditions: Rethinking HOG for All-in-One Image Restoration},
  author={Wu, Jiawei and Yang, Zhifei and Wang, Zhe and Jin, Zhi},
  booktitle={Proceedings of the AAAI Conference on Artificial Intelligence},
  year={2026}
}

@article{cui2026vivnet,
  title={Visual-in-Visual: A Unified and Efficient Baseline for Image Restoration},
  author={Cui, Yuning and Ren, Wenqi and Shi, Boxin and Knoll, Alois},
  journal={IEEE Transactions on Pattern Analysis and Machine Intelligence},
  volume={48},
  number={7},
  pages={7981--7999},
  year={2026}
}

@article{cai2016dehazenet,
  title={DehazeNet: An End-to-End System for Single Image Haze Removal},
  author={Cai, Bolun and Xu, Xiangmin and Jia, Kui and Qing, Chunmei and Tao, Dacheng},
  journal={IEEE Transactions on Image Processing},
  volume={25},
  number={11},
  pages={5187--5198},
  year={2016}
}

@inproceedings{li2017aodnet,
  title={AOD-Net: All-in-One Dehazing Network},
  author={Li, Boyi and Peng, Xiulian and Wang, Zhangyang and Xu, Jizheng and Feng, Dan},
  booktitle={Proceedings of the IEEE International Conference on Computer Vision},
  pages={4780--4788},
  year={2017}
}

@inproceedings{zhang2018dcpdn,
  title={Densely Connected Pyramid Dehazing Network},
  author={Zhang, He and Patel, Vishal M.},
  booktitle={Proceedings of the IEEE Conference on Computer Vision and Pattern Recognition},
  pages={3194--3203},
  year={2018}
}

@inproceedings{liu2019griddehazenet,
  title={GridDehazeNet: Attention-Based Multi-Scale Network for Image Dehazing},
  author={Liu, Xiaohong and Ma, Yongrui and Shi, Zhihao and Chen, Jun},
  booktitle={Proceedings of the IEEE/CVF International Conference on Computer Vision},
  pages={7314--7323},
  year={2019}
}

@inproceedings{qin2020ffa,
  title={FFA-Net: Feature Fusion Attention Network for Single Image Dehazing},
  author={Qin, Xu and Wang, Zhilin and Bai, Yuanchao and Xie, Xiaodong and Jia, Huizhu},
  booktitle={Proceedings of the AAAI Conference on Artificial Intelligence},
  volume={34},
  number={7},
  pages={11908--11915},
  year={2020}
}

@article{song2023dehazeformer,
  title={Vision Transformers for Single Image Dehazing},
  author={Song, Yuda and He, Zhuqing and Qian, Hui and Du, Xin},
  journal={IEEE Transactions on Image Processing},
  volume={32},
  pages={1927--1941},
  year={2023}
}

@inproceedings{wu2023ridcp,
  title={RIDCP: Revitalizing Real Image Dehazing via High-Quality Codebook Priors},
  author={Wu, Rui-Qi and Duan, Zheng-Peng and Guo, Chun-Le and Chai, Zhi and Li, Chong-Yi},
  booktitle={Proceedings of the IEEE/CVF Conference on Computer Vision and Pattern Recognition},
  pages={22282--22291},
  year={2023}
}

@article{liu2025funet,
  title={FUNet: Frequency-Aware and Uncertainty-Guiding Network for Rain-Hazy Image Restoration},
  author={Liu, Mengkun and Gao, Tao and Liu, Yao and Cao, Yuhan and Jiao, Licheng},
  journal={IEEE Transactions on Multimedia},
  volume={27},
  pages={9902--9917},
  year={2025}
}

@inproceedings{zheng2023curricular,
  title={Curricular Contrastive Regularization for Physics-Aware Single Image Dehazing},
  author={Zheng, Yu and Zhan, Jiahui and He, Shengfeng and Dong, Junyu and Du, Yong},
  booktitle={Proceedings of the IEEE/CVF Conference on Computer Vision and Pattern Recognition},
  pages={5785--5794},
  year={2023}
}

@inproceedings{cui2023irnext,
  title={IRNeXt: Rethinking Convolutional Network Design for Image Restoration},
  author={Cui, Yuning and Ren, Wenqi and Yang, Sining and Cao, Xiaochun and Knoll, Alois},
  booktitle={Proceedings of the 40th International Conference on Machine Learning},
  pages={6545--6564},
  year={2023}
}

@inproceedings{li2025mair,
  title={MaIR: A Locality- and Continuity-Preserving Mamba for Image Restoration},
  author={Li, Boyun and Zhao, Haiyu and Wang, Wenxin and Hu, Peng and Gou, Yuanbiao and Peng, Xi},
  booktitle={Proceedings of the IEEE/CVF Conference on Computer Vision and Pattern Recognition},
  year={2025}
}

@inproceedings{wu2026c2ssm,
  title={Scan Clusters, Not Pixels: A Cluster-Centric Paradigm for Efficient Ultra-High-Definition Image Restoration},
  author={Wu, Chen and Wang, Ling and Zheng, Zhuoran and Cui, Yuning and Yang, Zhixiong and Chen, Xiangyu and Zhang, Yue and Jiang, Weidong and Xia, Jingyuan},
  booktitle={Proceedings of the IEEE/CVF Conference on Computer Vision and Pattern Recognition},
  year={2026}
}

@inproceedings{dong2020msbdn,
  title={Multi-Scale Boosted Dehazing Network with Dense Feature Fusion},
  author={Dong, Hang and Pan, Jinshan and Xiang, Lei and Hu, Zhe and Zhang, Xinyi and Wang, Fei and Yang, Ming-Hsuan},
  booktitle={Proceedings of the IEEE/CVF Conference on Computer Vision and Pattern Recognition},
  pages={2157--2167},
  year={2020}
}

@inproceedings{zamir2022restormer,
  title={Restormer: Efficient Transformer for High-Resolution Image Restoration},
  author={Zamir, Syed Waqas and Arora, Aditya and Khan, Salman and Hayat, Munawar and Khan, Fahad Shahbaz and Yang, Ming-Hsuan},
  booktitle={Proceedings of the IEEE/CVF Conference on Computer Vision and Pattern Recognition},
  pages={5728--5739},
  year={2022}
}

@inproceedings{wang2022uformer,
  title={Uformer: A General U-Shaped Transformer for Image Restoration},
  author={Wang, Zhendong and Cun, Xiaodong and Bao, Jianmin and Zhou, Wengang and Liu, Jianzhuang and Li, Houqiang},
  booktitle={Proceedings of the IEEE/CVF Conference on Computer Vision and Pattern Recognition},
  pages={17683--17693},
  year={2022}
}

@inproceedings{potlapalli2023promptir,
  title={PromptIR: Prompting for All-in-One Blind Image Restoration},
  author={Potlapalli, Vaishnav and Zamir, Syed Waqas and Khan, Salman and Khan, Fahad Shahbaz},
  booktitle={Advances in Neural Information Processing Systems},
  volume={36},
  pages={71275--71293},
  year={2023}
}

@inproceedings{zheng2021wdc,
  title={Windowing Decomposition Convolutional Neural Network for Image Enhancement},
  author={Zheng, Chuanjun and Shi, Daming and Liu, Yukun},
  booktitle={Proceedings of the 29th ACM International Conference on Multimedia},
  pages={424--432},
  year={2021},
  doi={10.1145/3474085.3475181}
}

\end{document}